\documentclass{article}

\usepackage{iclr2018_conference,times}

\usepackage[utf8]{inputenc} % allow utf-8 input
\usepackage[T1]{fontenc}    % use 8-bit T1 fonts

\usepackage{hyperref,xcolor}       % hyperlinks

\definecolor{LinkColor}{rgb}{0,0,0}    
\hypersetup{colorlinks=true,
    linkcolor=LinkColor,
    citecolor=LinkColor,
   filecolor=LinkColor,
    menucolor=LinkColor,
    urlcolor=LinkColor}

\usepackage{url}            % simple URL typesetting
\usepackage{booktabs}       % professional-quality tables
\usepackage{amsfonts}       % blackboard math symbols
\usepackage{nicefrac}       % compact symbols for 1/2, etc.
\usepackage{microtype}      % microtypography

\usepackage{amsmath,amssymb,amsbsy,amstext,amsthm,mathrsfs}
\usepackage{color,soul}
\usepackage{algorithm}
\usepackage{subfigure}
\usepackage{latexsym,graphics,epsf,epsfig,psfrag,wrapfig}

\DeclareMathOperator*{\argmin}{arg\,min}

\newcommand{\bz}{\mathbf{z}}
\newcommand{\bx}{\mathbf{x}}

\newcommand{\bg}{\mathbf{g}}
\newcommand{\bff}{\mathbf{f}}

\newcommand{\by}{\mathbf{y}}

\newcommand{\bJ}{\mathbf{J}}

\newcommand{\bI}{\mathbf{I}}

\newcommand{\bG}{\mathbf{G}}

\newcommand{\bH}{\mathbf{H}}

\newcommand{\bv}{\mathbf{v}}
\newcommand{\bw}{\mathbf{w}}

\newcommand{\cR}{\mathcal{R}}

\newcommand{\cL}{\mathcal{L}}

\newcommand{\cS}{\mathcal{S}}

\title{Block-Diagonal Hessian-Free Optimization\\ for Training Neural Networks}
\author{Huishuai Zhang \thanks{This work was done when the author was an intern at Salesforce Research.} \\
Syracuse University\\
\texttt{hzhan23@syr.edu} \\
\And
\And
Caiming Xiong\thanks{Corresponding author},
James Bradbury \& Richard Socher \\
Salesforce Research \\
\texttt{\{cxiong, james.bradbury, rsocher\}@salesforce.com} 
}

\iclrfinalcopy

\begin{document}

\maketitle
% * <hzhan23@syr.edu> 2017-10-18T02:20:02.444Z:
%
% ^.

\begin{abstract}
Second-order methods for neural network optimization have several advantages over methods based on first-order gradient descent, including better scaling to large mini-batch sizes and fewer updates needed for convergence. But they are rarely applied to deep learning in practice because of high computational cost and the need for model-dependent algorithmic variations. We introduce a variant of the Hessian-free method that leverages a block-diagonal approximation of the generalized Gauss-Newton matrix. Our method computes the curvature approximation matrix only for pairs of parameters from the same layer or block of the neural network and performs conjugate gradient updates independently for each block. Experiments on deep autoencoders, deep convolutional networks, and multilayer LSTMs demonstrate better convergence and generalization compared to the original Hessian-free approach and the Adam method.
%TODO what I think I really want to say is better generalization to different model types compared to HF, and less generalization error compared to Adam with the same batch size

% Hessian-free (HF) optimization methods can use larger mini batches which is advantagrous in distributed training of large neural networks. However, they are more computationally expensive than first order methods like SGD due to many curvature-vector products required for each update.
% We introduce a model-dependent, block-wise HF method. Our method only computes blocks of the curvature approximation matrix for parameters that have high covariance \hl{high relation in the graph of network?}. Blocks are chosen specifically for each model architecture. In addition, our method uses smaller batches for estimating the curvature and larger batches for the first order update and objective function. This further reduces the computational cost.
% Theoretically we demonstrate that the blockwise HF method is equivalent to using a block diagonal curvature matrix. 
% Experiments on deep auto-encoders, deep convolutional networks and multilayer LSTMs demonstrate that our method improves both the speed and convergence of training compared to the original HF method.
%TODO make statement about how timing/convergence compares to Adam
\end{abstract}

\section{Introduction}\label{sec:introduction}
Deep neural networks have shown great success in computer vision \citep{he2016deep} and natural language processing tasks \citep{hochreiter1997long}.
These models are typically trained using first-order optimization methods like stochastic gradient descent (SGD) and its variants. The vanilla SGD does not incorporate any curvature information about the objective function, resulting in slow convergence in certain cases. Momentum \citep{qian1999momentum, nesterov2013introductory, sutskever2013importance} or adaptive gradient-based methods \citep{duchi2011adaptive, kingma2014adam} are sometimes used to rectify these issues.
These adaptive methods can be seen as implicitly computing finite-difference approximations to the diagonal entries of the Hessian matrix \citep{lecun1998efficient}.

A drawback of first-order methods in general, including adaptive ones, is that they perform best with small mini-batches \citep{dean2012large, zhang2015deep, das2016distributed, recht2011hogwild, chen2016revisiting}. This limits available parallelism and makes distributed training difficult. Moreover, in the distributed setting the gradients must be accumulated after each training update and the communication among workers may become a major bottleneck. The optimization methods that scale best to the distributed setting are those that can reach convergence with few parameter updates. The weakness of the first-order methods on this metric extends even to the convex case, where it was shown to be the result of correlation between the gradients at different data points in a mini-batch leading to ``overshooting'' in the direction of the correlation \citep{takac2013mini}. 

% Ideally, we can train with larger mini-batches so to increase parallelization and reduced overhead when updates are aggregated. However, there are two issues when using a large batch size. 
% The first is that a larger batch size still requires many parameter updates to get to the same convergence, even for convex case \citep{takac2013mini}. %The number of required iterations decreases linearly as enlarging mini-batch size only within small mini-batch regime. When the mini-batch size goes beyond a threshold, there is no benefit of increasing the mini-batch size in terms of reducing updates. 
% %TODO: If this was explained better, we could include it: This phenomenon is related to the correlation of data within a mini-batch. It contains more data points and induces more correlation within a large mini-batch, which hence hurts the speedup.  
% Secondly, a large batch size is shown to increase the generalization error by a considerable margin \citep{keskar2017large}. It is argued that this is due to the fact that large-batch methods tend to converge to sharp minima of the training function while small-batch methods converge to flat minima. 

In the case of deep neural networks, large mini-batch sizes lead to substantially increased generalization error \citep{keskar2017large, dinh2017sharp}. Although \citet{goyal2017accurate} recently successfully trained  deep ResNets on the ImageNet dataset in one hour with mini-batch size as large as 8192 by using the momentum-SGD equipped with some well-designed hyper-parameters, it also showed there is severe performance decay for even larger mini-batch sizes, which indeed corroborates the difficulty of training with large mini-batches. 
% Because the limiting factor in speeding up and parallelizing SGD-like methods is the small batch size, we focus on increasing the size in this paper. 
These difficulties motivate us to revisit the second-order optimization methods, which use the Hessian or other curvature matrices to rectify the gradient direction. The second-order methods employ more information about the local structure of the loss function, as they approximate it quadratically rather than linearly, and they scale better to large mini-batch sizes.

However, finding the exact minimum of a quadratic approximation to a loss function is infeasible in most deep neural networks because it involves inverting an $N$-by-$N$ curvature matrix for a parameter count of $N$. The Hessian-free (HF) methods \citep{martens2010deep, martens2011learning,byrd2011use, byrd2012sample} minimize the quadratic function that locally approximates the loss using the conjugate gradient (CG) method instead. This involves evaluating a sequence of curvature-vector products rather than explicitly inverting---or even computing---the curvature matrix or Hessian. The Hessian-vector product can be calculated efficiently using one forward pass and one backward pass \citep{pearlmutter1994fast}, while other curvature-vector products have similarly efficient algorithms \citep{schraudolph2002fast,martens2012training}.

Normally, the HF method requires many hundreds of CG iterations for one update, which makes even a single optimization step fairly computationally expensive. Thus,  when comparing HF to first-order methods, the benefit in terms of fewer iterations from incorporating curvature information often does not compensate for the added computational burden.

We propose using a block-diagonal approximation to the curvature matrix to improve Hessian-free convergence properties, inspired by several results that link these two concepts for other optimization methods. \cite{collobert2004large} argues that when training a multilayer perceptron (MLP) with one hidden layer the gradient descent  converges faster with the cross-entropy loss than with mean squared error because its Hessian is more closely block-diagonal. A block-diagonal approximation of the Fisher information matrix, one kind of curvature matrix, has also been shown to improve the performance of the online natural gradient method \citep{roux2008topmoumoute} for training a one-layer MLP.

The advantage of a block-diagonal Hessian-free method is that updates to certain subsets of the parameters are independent of the gradients for other subsets. This makes the subproblem separable and reduces the complexity of the local search space \citep{collobert2004large}. We hypothesize that using the block-diagonal approximation of the curvature matrix may make the Hessian-free method more robust to noise that results from using a relatively small mini-batch for curvature estimation.

In the cases of \cite{collobert2004large} and \cite{roux2008topmoumoute}, the parameter blocks for which the Hessian or Fisher matrix is block-diagonal are composed of all weights and biases involved in computing the activation of each neuron in the hidden and output layers. Thus it equates to the statement that gradient interactions among weights that affect a single output neuron are greater than those between weights that affect two different neurons.

In order to strike a balance between the curvature information provided by additional Hessian terms and the potential benefits of a more nearly block-diagonal curvature matrix, and adapt the concept to more complex contemporary neural network models, we choose to treat each layer or submodule of a deep neural network as a parameter block instead. Thus, different from \cite{collobert2004large} and \cite{roux2008topmoumoute}, our hypothesis then becomes that gradient interactions among weights in a single layer are more useful for training than those between weights in different layers.

We now introduce our block-diagonal Hessian-free method in detail, then test this hypothesis by comparing the performance of our method on a deep autoencoder, a deep convolutional network, and a multilayer LSTM to the original Hessian-free method \citep{martens2016second} and the Adam method \citep{kingma2014adam}.

% % In order to use small mini-batches for curvature estimation while still benefiting from the HF method, we propose a model-dependent, blockwise Hessian free (blockwise-HF) method. The blockwise-HF method uses block diagonal information of the curvature matrix. 
% %TODO This Hence, ... is not obvious here yet.
% % Hence it is more robust to noise induced by small curvature mini-batch than the HF method. 
% The idea behind blockwise-HF is that deep neural networks are built of layers and blocks, and thus it is reasonable to assume that inter correlation between different blocks is less important than intra correlation within each block. 
% %TODO I guess that's what we need to verify still..? if it's true, we should add a plot
% By using the blockwise-HF method, we explicitly ignore the curvature information between different blocks. 
% %TODO this should be given some evidence.
% %and remedy the affect of inaccurate curvature matrix due to small curvature mini-batch. 
\section{The Block-Diagonal Hessian-Free Method}\label{sec:algorithm}
In this section, we describe the block-diagonal HF method in detail and compare it with the original HF method  \citep{martens2010deep, martens2011learning}.
% The HF method relies on several modifications to work well in practice. We describe how these tricks apply to the block-diagonal HF method and discuss their properties.

Throughout the paper, we use boldface lowercase letters to denote column vectors, boldface capital letters to denote matrices or tensors, and the superscript $^\top$ to denote the transpose. We denote an input sample and its label as $(\bx, \by)$, the output of the network as $\bff(\bx, \bw)$, and the loss as $\ell(\by, \bff(\bx, \bw))$, where $\bw$ refers to the network parameters flattened to a single vector.

%Background on the 
%\subsection{Hessian-Free Method}\label{subsec:HF}
\subsection{The Block-Diagonal Hessian-Free Method}\label{subsec:blockwise}
We first recall how second-order optimization works. For each parameter update, the second-order method finds $\Delta \bw$ that minimizes a local quadratic approximation $q(\bw+\Delta \bw)$ of the objective function $\ell(\cdot)$ at point $\bw$:
\begin{flalign}
q(\bw+\Delta \bw) := \ell(\bw) + \Delta \bw^\top \nabla \ell(\bw) + \frac{1}{2} \Delta \bw^\top \bG(\bw)\Delta \bw, \label{eq:quadratic_approximation}
\end{flalign}
where $\bG(\bw)$ is some curvature matrix of $\ell(\cdot)$ at $\bw$, such as the Hessian matrix or the generalized Gauss-Newton matrix \citep{martens2012training}. The resulting sub-problem of 
\begin{flalign}
\argmin_{\Delta \bw} \Delta\bw^\top \nabla \ell +\frac{1}{2} \Delta\bw^\top \bG \Delta\bw \label{eq:subopt}
\end{flalign}
is solved using conjugate gradient (CG), a procedure that only requires evaluating a series of matrix-vector products $\bG\bv$. 

There exist efficient algorithms for computing these matrix-vector products given a computation-graph representation of the loss function. If the curvature matrix $\bG$ is the Hessian matrix, \eqref{eq:quadratic_approximation} is the second-order Taylor expansion and the Hessian-vector product can be computed as the gradient of the directional derivative of the loss function in the direction of $\bv$, operations also known as the L- and R-operators $\cL\{\cdot\}$ and $\cR_\bv\{\cdot\}$ respectively:
\begin{flalign}
\bH \bv = \frac{\partial^2 \ell}{\partial^2 \bw} \bv = \nabla_\bw(\bv^\top\nabla_\bw\ell) = \cL\{\cR_\bv\{\ell(\bw)\}\}.
\end{flalign}
The R-operator can be implemented as a single forward traversal of the computation graph (applying forward-mode automatic differentiation), while the L-operator requires a backward traversal (reverse-mode automatic differentiation) \citep{pearlmutter1994fast, baydin2015automatic}. The Hessian-vector product can also be computed as the gradient of the dot product of a vector and the gradient; that method does not require the R-operator but has twice the computational cost.

However, the objective of deep neural networks is non-convex and the Hessian matrix may have a mixture of positive and negative eigenvalues, which makes the optimization problem \eqref{eq:subopt} unstable. It is common to use the generalized Gauss-Newton matrix \citep{schraudolph2002fast} as a substitute curvature matrix, as it is always positive semidefinite if the objective function can be expressed as the composition of two functions $\ell(\bff(\bw))$ with $\ell$ convex, a property satisfied by most training objectives. For a curvature mini-batch of data $\cS_c$, the generalized Gauss-Newton matrix is defined as
\begin{flalign}
\bG := \frac{1}{|\cS_c|}\sum_{(\bx, \by)\in \cS_c} \bJ^\top\bH_{\ell} \bJ,
\end{flalign}
where $\bJ$ is the Jacobian matrix of derivatives of network outputs with respect to the parameters $\bJ:=\frac{\partial \bff}{\partial \bw}$ and $\bH_{\ell}$ is the Hessian matrix of the objective with respect to the network outputs $\bH_{\ell}: =\frac{\partial^2 \ell}{\partial^2 \bff}$. It is an approximation to the Hessian that results from dropping terms that involve second derivatives of $\bff$ \citep{martens2012training}.

The Gauss-Newton vector product $\bG\bv$ can also be evaluated as
\begin{flalign}
\bG\bv =
(\bJ^\top\bH_{\ell} \bJ) \bv =
\bJ^\top \left(\bH_{\ell}\left(\bJ \bv\right)\right).
%=\cL_{\cL\{\cR_{\cR_\bv\{\bff(\bw)\}}\{\ell(\bff)\}\}} \{\bff(\bw)\}.
\end{flalign}
In an automatic differentiation package like Theano \citep{al2016theano}, this requires one forward-mode and one reverse-mode traversal of the computation graphs of each of $\ell(\bff)$ and $\bff(\bw)$.

%\subsection{The Block-Diagonal Hessian-Free Method}\label{subsec:blockwise}

However, it is still  inefficient to solve problem \eqref{eq:subopt}  for a deep neural network with a large number of parameters, so we propose the block-diagonal Hessian-free method. We first split the network parameters into a set of parameter blocks. For instance, each block may contain the parameters from one layer or a group of adjacent layers. Then the sub-problems corresponding to each block are solved separately, while their solutions are concatenated together to produce a single update. 

Specifically, if there are $B$ blocks in total, the parameter vector can be rewritten as $\bw=[\bw_{(1)}; \bw_{(2)}; \ldots; \bw_{(B)}]$. Similarly, we split the gradient into blocks as  $\nabla \ell(\bw) = [\nabla_{(1)} \ell; \nabla_{(2)}\ell; \ldots; \nabla_{(B)}\ell]$, where $\nabla_{(b)}\ell$ is the vector that contains the gradient only with respect to the parameters in block $b$. We further split the curvature matrix into $B\times B$ square blocks and let $\bG_{(b)}$ be the $b$-th diagonal block of $\bG$. Then we obtain separate sub-problems for each block as follows: 
 \begin{flalign*}
 &\argmin_{\Delta \bw_{(1)}}  \Delta \bw_{(1)}^\top \nabla_{(1)} \ell+ \frac{1}{2} \Delta \bw_{(1)}^\top \bG_{(1)}\Delta \bw_{(1)},\\
  &\argmin_{\Delta \bw_{(2)}} \Delta \bw_{(2)}^\top \nabla_{(2)} \ell+ \frac{1}{2} \Delta \bw_{(2)}^\top \bG_{(2)}\Delta \bw_{(2)},\\
 &\ldots,\\
 &\argmin_{\Delta \bw_{(B)}} \Delta \bw_{(B)}^\top \nabla_{(B)} \ell+ \frac{1}{2} \Delta \bw_{(B)}^\top \bG_{(B)}\Delta \bw_{(B)}.
 \end{flalign*}
 We solve these sub-problems separately by conjugate gradient and concatenate their solutions together. Hence $\Delta \bw =[\Delta\bw_{(1)}; \ldots; \Delta\bw_{(B)}]$ will be our update (see Algorithm \ref{alg:blockwise}).

\begin{algorithm}[t!]
\caption{Block-Diagonal Hessian-Free Method}\label{alg:blockwise}
\textbf{Input}: Training data set $\cS_T=\{(\bx_i,\by_i), i=1, \ldots, |\cS_T|\}$; Neural network output function $\bz_i = \bff(\bx_i, \bw)$ with parameters $\bw$ and loss function $\ell(\bz_i, \by_i)$; Hyper-parameters: maximum loops \texttt{max\_loops}, maximum conjugate gradient iterations \texttt{max\_cg\_iters}, CG stop criterion \texttt{cg\_stop\_criterion}, learning rate $\alpha$\\
\textbf{Block partition}: Partition the network parameters into $B$ blocks, i.e., $\bw=[\bw_{(1)}; \bw_{(2)}; \ldots; \bw_{(B)}]$

 \textbf{For $k=1, \ldots,$ \texttt{max\_loops}}:
	\begin{enumerate}
	\item Choose a gradient mini-batch $\cS_g\subset \cS_T$ to calculate the gradient 
$
    	\bg = [\bg_{(1)}; \ldots; \bg_{(B)}]
$
    \item Choose a curvature mini-batch $\cS_c \subset \cS_g$ to calculate the curvature-vector product
    \item CG iterations:\\
    	For $b=1, \ldots, B$, solve
$
\argmin_{\Delta \bw_{(b)}} \Delta \bw_{(b)}^\top \nabla_{(b)} \ell+ \frac{1}{2} \Delta \bw_{(b)}^\top \bG_{(b)}\Delta \bw_{(b)}
$
        by CG with \texttt{max\_cg\_iters} and \texttt{cg\_stop\_criterion}. These suboptimizations may be performed in parallel.
     \item Aggregate $\Delta \bw\leftarrow[\Delta \bw_{(1)}; \ldots; \Delta \bw_{(B)}]$ and update $\bw \leftarrow \bw + \alpha\Delta \bw$
	\end{enumerate}
\vspace{-5pt}
\end{algorithm}

The $b$-th sub-problem of the block-diagonal HF method is equivalent to minimizing the overall objective \eqref{eq:quadratic_approximation} with constraint $\Delta \bw_j=0$ for $j\notin (b)$, since the second-order term of such a constrained objective is zero for all terms in $\bG$ not in $\bG_{(b)}$.
This confirms that block-diagonal HF as described above is equivalent to ordinary HF with the curvature matrix replaced by a block-diagonal approximation that includes only terms involving pairs of parameters from the same block.

The problem  \eqref{eq:subopt} has been separated into independent sub-problems for each block, reducing the dimensionality of the search space that CG needs to consider. Although we have $B$ sub-problems to solve for one update, each sub-problem has smaller size and requires fewer CG iterations. Hence, the total compute needs are on par with those of the HF method with the same mini-batch sizes; if the independent sub-problems can be executed in parallel (e.g., on multiple nodes in a distributed system), there is potential for up to $B$-fold speed improvement.
As we demonstrate below, block-diagonal Hessian-free achieves better performance than the HF method on deep autoencoders, multilayer LSTMs, and deep CNNs.

\subsection{Implementation Details}\label{subsec:implementation}

%Each iterative optimization algorithm is to find a path that lead to the optimum of the objective function. Provide the parameter update that decrease the objective most. Gradient descent follows the rule that each parameter update is on the gradient direction which is the steepest descent direction. However, the steepest direction is only within a extremely small ball, gradient is steepest descent direction in limitation. The direction is steepest descent only if you take infinite small steps.  However to make the gradient descent, we cannot take infinite small step size. If we enlarge the ball we search over, we can see the gradient direction is no longer the steepest descent direction. It we take large step size, the gradient descent often moves to the optimum with oscillating path. Thus if we are allowed to access curvature information beyond the gradient, we can amend the gradient by multiplying with the inverse of the curvature matrix. This is the second-order method, which often gives us faster convergence.  However, in large system, it is prohibitive to evaluate and store the Hessian matrix (if there are $n$ parameters in total, there are $n^2$ elements in Hessian matrix). 
% The blockwise HF method is described in Algorithm \ref{alg:blockwise} and we now discuss several implementation details.

We partition the network parameters into blocks based on the architecture of the network. %Although deep neural networks may have arbitrary structures, they are built from simple layers or other building  blocks. 
When partitioning the network parameters, we try to define roughly equal sized blocks. This allows each sub-problem to make roughly similar progress with the same number of CG iterations.
We seek to partition the network such that parameters whose gradients we expect to be strongly correlated are part of the same block.
% Optimal parameter partitions are still an open problem and depend on intuition. While we do not have a thorough evaluation on how to obtain the best partitions, it turns out that intuitive partitions work well in practice. 
For example, in our experiment we split the autoencoder network into two blocks: one for the encoder and one for the decoder.
For the multilayer LSTM, we treat each layer of recurrent cells as a block.
And for the deep CNN, we divide the convolutional layers into three contiguous blocks.
%{\color{red}Should we do some experiments about how fragile the result will be for different partitions.}
% Yes, and if we get results, we should update this paragraph 

When solving the problem \eqref{eq:subopt}, we use truncated conjugate gradient \citep{yuan2000truncated}. This means we terminate the CG iteration before finding the local minimum. There are two reasons to do this truncation. First, CG iterations are expensive and later iterations of CG provide diminishing improvements. More importantly, when we use mini-batches to evaluate the curvature-vector product, early termination of CG keeps the update from overfitting to the specific mini-batch.
% maybe rework this: Strong truncated CG makes the HF-style methods works stably in the early stage of the training (second-order method is not stable if the iterates are far from optimal points). We set a maximum number of CG iterations to control the CG truncation. 
%We can also construct criteria to monitor the progress of CG and terminate it adaptively.
%define criteria here or mention it later together with these criteria or if we dont do it, dont mention it.

One way to reduce the computational burden of the HF method is to use smaller mini-batch sizes to evaluate the curvature-vector product while still using a large mini-batch to evaluate the objective and the gradient \citep{byrd2011use, byrd2012sample, kiros2013training}. \cite{martens2010deep} similarly implements the HF method using the full dataset to evaluate the objective and the gradient, and mini-batches to calculate the curvature-vector products. This is possible because Newton-like methods are more tolerant to approximations of the Hessian than they are to that of the gradient \citep{byrd2011use}. In our implementation, the curvature mini-batch is chosen to be a strict subset of the gradient mini-batch as shown in Algorithm \ref{alg:blockwise}.

However, small mini-batches inevitably make the curvature estimation deviate from the true curvature, reducing the convergence benefits of the HF method over first-order optimization \citep{martens2012training}. In practice it is not trivial to choose a mini-batch size that balances accurate estimation of curvature and the computational burden \citep{byrd2011use}. The key to making Hessian-free methods, including block-diagonal Hessian-free, converge well with small curvature mini-batches is to use short CG runs to tackle mini-batch overfitting.
% This is because when using small mini-batches, the curvature matrix differs significantly from the true curvature. Since more CG iterations overfit the curvature estimate of the mini-batch and harm the optimization, we suggest that smaller curvature mini-batch sizes have even fewer CG runs.
 
\cite{martens2010deep} suggests using factored Tikhonov damping to make the HF method more stable. With damping, $\hat{\bG}:=\bG+d\bI$ is used as the curvature matrix to make the curvature ``more'' positive definite, where $d$ is the intensity of damping. %Moreover, \cite{martens2010deep} uses Tikhonov damping schedule that adapts the damping intensity $d$ according to some optimization criterion.
We also incorporate damping in many of our experiments. For the sake of comparison, we use the same damping strength for the HF method and the block-diagonal HF method and choose a fixed value for each experiment.

Another suggestion made by \citep{martens2010deep} is to use a form of ``momentum'' to accelerate the HF method. Here, momentum means initializing the CG algorithm with the last CG solution scaled by some constant close to 1, rather than initializing it randomly or to the zero vector. This change often brings additional speedup with little extra computation. We apply a fixed momentum value of $0.95$ for all experiments.

%\cite{martens2010deep, martens2011learning} also introduce other tricks to improve the HF method, such as CG backtracking and preconditioning, as well as adaptive hyperparameter scheduling.   \cite{wiesler2013investigations} investigates some of these variations and concludes that CG preconditioning and CG backtracking techniques are not very helpful in reducing the number of CG iterations needed but do increase the computational burden. To keep the comparison simple, we do not adopt these in our experiments.

We also adopt fixed hyper-parameter settings across the experiments, rather than an adaptive schedule. One reason is that the statistics that control the adaptive hyper-parameter scheduling can cost more than the gradient and curvature-vector product evaluation, which makes the HF method even slower. %We test the LM heuristic for adjusting the damping parameter and find that the LM adjusting rules of achieving the best performance are different from one experiment to another. Moreover, a fixed (Tikhonov) damping parameter achieves almost the same performance in our experiments, which also avoids the trouble of choosing the LM adjusting rule.  
Furthermore, these tricks are not independent and it is often unclear how to adjust and fit them to every scenario. Our fixed hyperparameters work well in practice across the three different neural network architectures we investigated.

%The HF method incorporates the curvature information in the traditional forward-backward fashion. After backward step, we have gradients for each parameter of the network. Instead of doing a normal gradient descent step, we minimize a quadratic function (approximate the original loss function locally) by conjugate gradient method.

%Newton-like methods are very tolerant to the choice of Hessian and can make good use of limited curvature information. Specifically, if $B$ is any symmetric and positive definite matrix and if we apply any number of CG steps to the system $Bd = - \nabla J(w_k)$, the resulting Newton-CG step is a descent direction for $J(w_k)$. In machine learning applications, if the loss funtion is convex then $\nabla^2 J_S$ will be positive semidefinite for any nonempty choice of $S$. 

%%%%%%%%%%%%%%%%%%%%%%%%%%%%%%%

\section{Related Work}\label{subsec:related}
%A variety of second-order optimization techniques have been applied to nonlinear functions such as neural networks.

%One line of research has used the truncated conjugate gradient method to minimize non-linear functions, e.g.,  convex quadratics \citep{yuan2000truncated} and convex functions useful in machine learning applications \cite{byrd2011use, byrd2012sample}. In the field of deep learning, \cite{martens2010deep, martens2011learning} have studied the Hessian-free (HF) method, an approach based on truncated conjugate gradient, to train deep neural networks including deep autoencoders and RNNs. More recently, 
% The Hessian-free optimization method of \cite{martens2010deep} has recently been applied successfully to train multilayer LSTMs \cite{krause2015optimizing, cho2015hessian} and to train deep autoencoders in a distributed setting \citep{he2016large}.

The Hessian matrix is indefinite for nonconvex objectives, which makes the second-order method unstable as the local quadratic approximation becomes unbounded from below.  \citep{martens2012training} advocates using the generalized Gauss-Newton matrix \citep{schraudolph2002fast} as the curvature matrix instead, which is guaranteed to be positive semi-definite. Another way to circumvent the indefiniteness of the Hessian is to use the Fisher information matrix as the curvature matrix; this approach has been widely studied under the name ``natural gradient descent'' \citep{amari2007methods, amari1998natural, pascanu2013revisiting, roux2008topmoumoute}. In some cases these two curvature matrices are exactly equivalent \citep{pascanu2013revisiting, martens2016second}. It has also been argued that the negative eigenvalues of the full Hessian are helpful for finding parameters with lower energy, e.g., in the saddle-free Newton method \citep{dauphin2014identifying} and in an approach that mixes the Hessian and Gauss-Newton matrices \citep{he2016large}.

%\citep{ngiam2011optimization}

%\cite{collobert2004large} demonstrates that the Hessian of an MLP with one hidden layer and the cross entropy objective converges to a block-diagonal matrix during first-order optimization, where the blocks are composed of the weights linking to each output unit in both hidden layer and output layer. A block-diagonal approximation of Fisher matrix has also been shown to improve the performance of the online natural gradient method  \citep{roux2008topmoumoute} for training a similar MLP.

Recently, \cite{martens2015optimizing, grosse2016kronecker}, and \cite{ba2017distributed} propose the K-FAC method to approximate the natural gradient using a block-diagonal or block-tridiagonal approximation to the \textit{inverse} of the Fisher information matrix, and demonstrate the advantages over first-order methods of a specialized version of this optimizer tailored to deep convolutional networks. In their work, the parameters are partitioned into blocks of similar size and structure to those used in our method.

\section{Experiments}

We evaluate the performance of the block-diagonal HF method on three deep architectures: a deep autoencoder on the MNIST dataset, a 3-layer LSTM for downsampled sequential MNIST classification, and a deep CNN based on the ResNet architecture for CIFAR10 classification.
For all three experiments, we first compare the performance of the block-diagonal HF method with that of Adam \citep{kingma2014adam} to demonstrate that block-diagonal Hessian-free is able to handle large batch size more efficiently. We then demonstrate the advantage of the block-diagonal method over ordinary Hessian-free by comparing their performance at various curvature mini-batch sizes.

Although the block-diagonal HF method needs to solve more  quadratic minimization problems, each sub-problem is much smaller and the computation time is similar to the HF method. %In the following experiments, we observe that block-diagonal HF takes 10\%-15\% more time than the HF method with the same hyperparameter settings. 
We note that the independence of the CG sub-problems means the block-diagonal method is particularly amenable to a distributed implementation. 
%Martens advocates to use exponentially increasing batch-size to achieve better accuracy. We have applied this suggestion and found it helps to improve the training loss and test accuracy a little bit in the later stage of training process. In order to keep the comparison simple, we fix the batch-size throughout the training process.

 %However, these stochastic HF methods often do not perform as good as the original HF using large curvature-vector batches and large number of CG iterations. 

We use the Lasagne deep learning framework \citep{lasagne} based on Theano \citep{al2016theano} for our implementation of the HF and block-diagonal HF methods, as we found no other software framework to support both convenient definition of deep neural networks and the forward-mode automatic differentiation required to implement the R-operator.

\subsection{Deep Autoencoder}

\begin{figure}[t!]
\centering 
% \subfigure{
% \includegraphics[width=2in]{images/AEtrain.pdf}}
% \subfigure{
% \includegraphics[width=2in]{images/AEtest.pdf}}
% \subfigure{
% \includegraphics[width=2in]{images/AEtrainhf.pdf}}
% \subfigure{
% \includegraphics[width=2in]{images/AEtesthf.pdf}}
\includegraphics[width=5.0in]{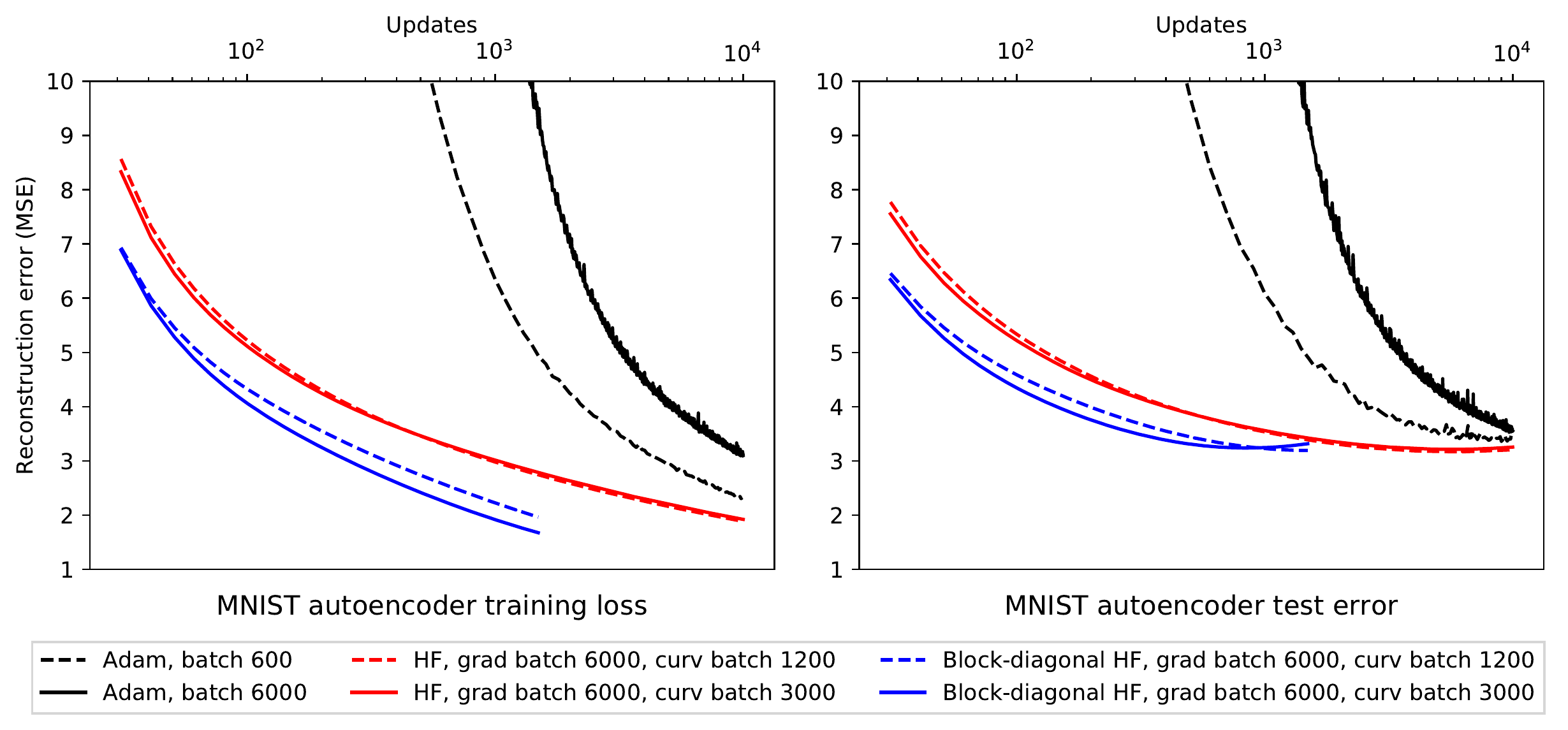}
\vspace{-10pt}
\caption{Performance comparison for a deep feedforward autoencoder on MNIST. The early epochs of Adam have reconstruction error greater than 10, while all models are run until the test error increases or for 1000 epochs.
%The upper two plots compare Adam (with mini-batch sizes 600 and 6000) to block-diagonal HF with gradient mini-batch size 6000 and curvature mini-batch size 3000. The lower two compare block-diagonal HF to HF, with gradient mini-batch size 6000 and curvature mini-batch sizes 3000 and 1200. All plots use a log scale for the number of updates.
}
\label{fig:autoencoder}
\vspace{-15pt}
\end{figure}

Our first experiment is conducted on a deep autoencoder task. The goal of a neural network autoencoder is to learn a low-dimensional representation (encoding) of data from an input distribution. The ``encoder'' part, a multi-layer feedforward network, maps the input data to a low-dimensional vector representation while the ``decoder'' part, another multi-layer feedforward network, reconstructs the input data given the low-dimensional vector representation. The autoencoder is trained by minimizing the reconstruction error.

The MNIST dataset \citep{mnist} is composed of handwritten digits of size $28 \times 28$ with $60,000$ training samples and $10,000$ test samples. The pixel values of both the training and test data are rescaled to $[0,1]$.

Our autoencoder is composed of an encoder with three hidden layers and state sizes 784-1000-500-250-30, followed by a decoder that is the mirror image of the encoder\footnote{ The model of autoencoder is the same as that in \cite{hinton2006fast} and \cite{martens2010deep} for easy comparison.}. We use the $\tanh$ activation function and the mean squared error loss function.

For hyperparameters, we use a fixed learning rate of $0.1$, no damping, and maximum CG iterations \texttt{max\_cg\_iters} $= 30$ for both the HF and block-diagonal HF methods. For block-diagonal HF, we define two blocks: one block for the encoder and the other for the decoder. For Adam, we use the default setting in Lasagne with learning rate 0.001, $\beta_1$=0.9, $\beta_2$=0.999, and $\epsilon=1\times 10^{-8}$.

A performance comparison between Adam, HF, and block-diagonal HF is shown in Figure \ref{fig:autoencoder}. For Adam, the number of dataset epochs needed to converge and the final achievable reconstruction error are heavily affected by the mini-batch size, with a similar number of updates required for small-mini-batch and large-mini-batch training. Our block-diagonal HF method with large mini-batch size achieves approximately the same reconstruction error as Adam with small mini-batches while requiring an order of magnitude fewer updates to converge compared to Adam with either small or large mini-batches.  Moreover, block-diagonal Hessian-free provides consistently better reconstruction error---on both the train and test sets---than the HF method over the entire course of training. This advantage holds across different values of the curvature mini-batch size.

%We also provide the run time per epoch. For the maximum number of CGs, the blockwise HF method consumes $10\%$ more time than the HF method.
%conclusion.

% \begin{table}[t]
%   \caption{GPU time consuming per epoch, fix CGs 30}
%   \label{tab:aetime}
%   \centering
%   \begin{tabular}{l|lll}%{lll|ll|ll}
%     \toprule
%    	 Curvature mini-batch size & 600 & 1200 & 3000\\
%         \midrule
%       The HF method (s) & 9.9 & 12.8 & 22.8\\
%    	\midrule
% The blockwise HF method (s)  & 10.7 & 14.4 & 25.5\\
%          \bottomrule
%   \end{tabular}
% \end{table}

% \begin{figure}[th]
% \centering 
% \includegraphics[width=2.5in]{images/AEtrainhf.pdf}
% \includegraphics[width=2.5in]{images/AEtesthf.pdf}
% \caption{Performance comparison on autoencoder: The y-axis is the reconstruction error and x-axis is the epochs in log scale.}
% \label{fig:ae30cgs}
% \end{figure}
\vspace{-10pt}
\subsection{Multilayer LSTM}

Our second experiment is conducted using a three-layer stacked LSTM on the sequential MNIST classification task. The MNIST data $(28\times28)$ is downsampled to $(7\times7)$ by average pooling. The neural network has three LSTM \citep{hochreiter1997long, gers2002learning} layers followed by a fully-connected layer on the final layer's last hidden state. Each LSTM has 10 hidden units with peephole connections \citep{gers2002learning}.
% The last LSTM layer outputs the final hidden state and the other two LSTM layers output the hidden state sequences.

For HF and block-diagonal HF, we use a fixed learning rate of $0.1$, damping strength $0.01$, and maximum CG iterations \texttt{max\_cg\_iter} $= 100$. The block-diagonal method has three blocks---one block for each LSTM layer, with the top block also containing the fully-connected layer. For Adam, we again use a learning rate of 0.001, $\beta_1$=0.9, $\beta_2$=0.999, and $\epsilon=1\times 10^{-8}$.

% We apply the CG stop criteria for the HF and blockwise HF method.
%We compare three algorithms Adam \citep{kingma2014adam}, the HF method \citep{martens2012training} and the blockwise HF method. The Adam method is applied with default hyper-parameter setting with $lr=0.001, \beta1=0.9, \beta2=0.999, \epsilon=1e-08$. We use the same hyper-parameter settings for the HF method and the blockwise HF method, $lr=0.1, \text{damping}=0.01, \text{MAX\_ITERS}=100, \epsilon=1e-3$. The gradient mini-batch size is 6000. We use several curvature mini-batch sizes 3000, 1200, 600 and compare their performances. 
\begin{figure}[t!]
\centering 
% \subfigure{
% \includegraphics[width=2in]{images/lstm6000train.pdf}}
% \subfigure{
% \includegraphics[width=2in]{images/lstm6000test.pdf}}
% \subfigure{
% \includegraphics[width=2in]{images/lstm6000trainhf.pdf}}
% \subfigure{
% \includegraphics[width=2in]{images/lstm6000testhf.pdf}}
\vspace{-15pt}
\includegraphics[width=5.0in]{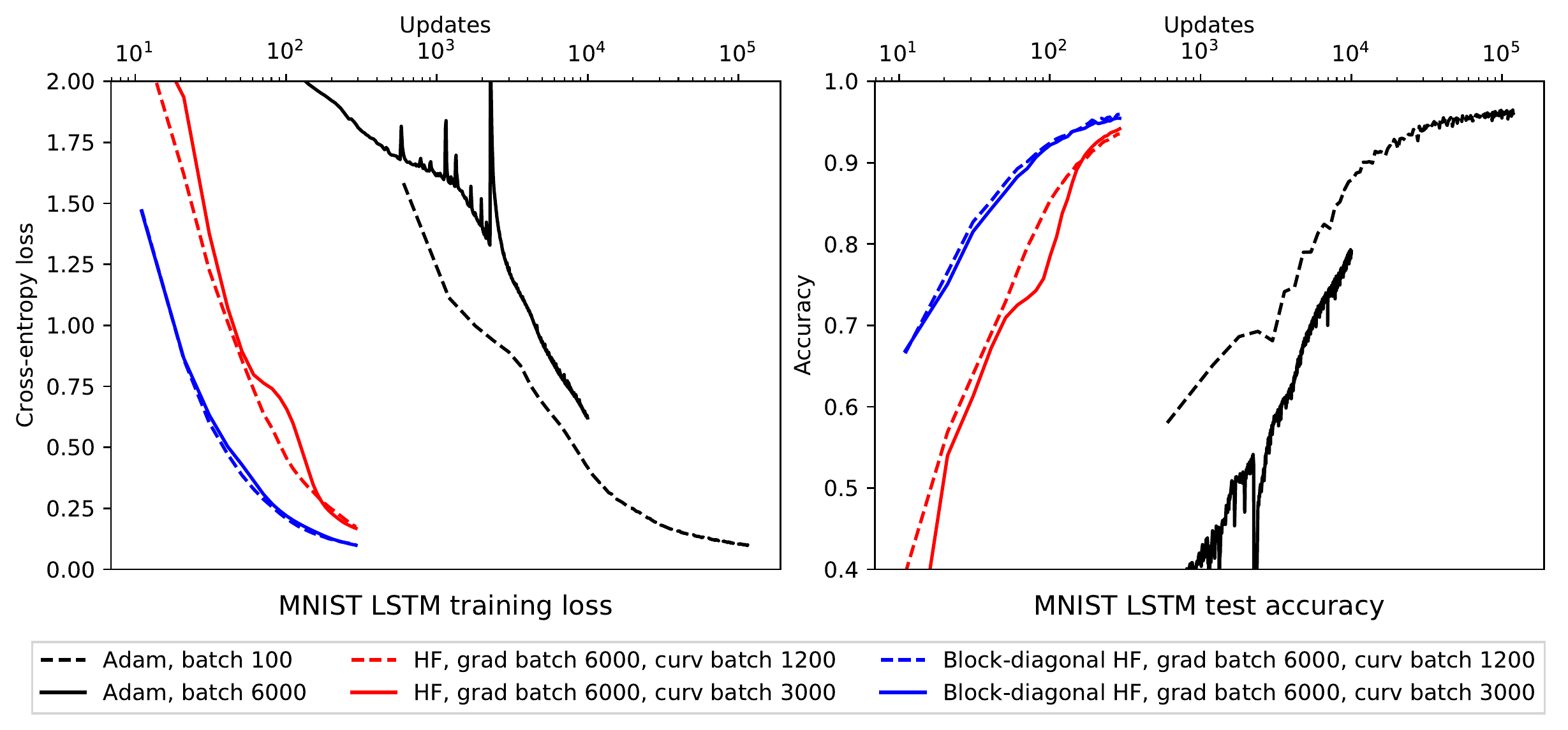}
\caption{Performance comparison for a 3-layer stacked LSTM on the sequential MNIST $(7\times7)$ classification task. The early epochs of Adam have loss greater than 2, while all models are run until the test accuracy decreases or for 1000 epochs.
%The upper two plots compare Adam (with mini-batch sizes 6000, 2000, and 100) to block-diagonal HF (with gradient mini-batch 6000 and curvature mini-batch 1200). The lower two compare block-diagonal HF to HF, both with gradient mini-batch size 6000 and curvature mini-batch sizes 3000, 1200, and 600. All plots use a log scale for the number of updates.
}
\label{fig:lstm}
\vspace{-20pt}
\end{figure}

A performance comparison between block-diagonal HF, HF, and Adam is found in Figure \ref{fig:lstm}. Similar to the autoencoder case, the block-diagonal method with large mini-batches requires far fewer updates to achieve lower training loss and better test accuracy than Adam with any mini-batch size. Furthermore, compared to HF, the block-diagonal HF method requires fewer updates, achieves better minima, and exhibits less performance deterioration for small curvature mini-batch sizes. %it achieves the same   It can be seen that the convergence rate and final achievable reconstruction error of Adam are heavily affected by the mini-batch size. The blockwise HF method with very large batch size achieves almost the same reconstruction error as Adam with small batch size and requires order-level fewer epochs to converge compared to Adam with the same large mini-batch size.  Moreover, The blockwise HF method exhibits consistent advantage over the HF method on both training and test sets.

% \begin{figure}[h]
% \centering 
% \includegraphics[width=2.5in]{images/lstm6000trainhf.pdf}
% \includegraphics[width=2.5in]{images/lstm6000testhf.pdf}
% \caption{Performance comparison on 3-layer LSTM network for sequential MNIST $(7\times7)$ classification task: Blockwise HF  vs. HF. The gradient batch size is $6000$ and curvature mini-batch sizes are $3000, 1200, 600$ for both methods.}
% \label{fig:lstmhf}
% \end{figure}

\subsection{Deep Convolutional Neural Network}
We also train a deep convolutional neural network (CNN) for the CIFAR-10 classification task with the three optimization methods. The CIFAR-10 dataset has $50,000$ training samples and $10,000$ test samples, and each sample is a $32 \times 32$ image with three channels.

Our model is a simplified version of the ResNet architecture \citep{he2016deep}. It has one convolutional layer $(16\times 32 \times 32)$ at the bottom followed by three residual blocks and a fully-connected layer at the top. We did not include batch normalization layers\footnote{ Computing the Hessian-vector product becomes extremely slow when involving the batch normalization layers with the Theano framework.}.
% More implementation detail can be found in Recipes of Lasagne \citep{lasagne}.

For HF and block-diagonal HF, we use a fixed learning rate of $0.1$, damping strength $0.1$, and maximum CG iterations \texttt{max\_cg\_iter} $= 30$. The block-diagonal method again has three blocks---one for each residual block, with the top and bottom blocks also containing the fully-connected and convolution layers respectively. We use the same default Adam hyperparameters.

%It is shown that decaying the learning rate is critical for residual network achieving good performance \citep{he2016deep}  and \cite{grosse2016kronecker} suggests that Polyak averaging \citep{polyak1992acceleration} can obviate the need of scheduling the learning rate. To make the comparison simple, we apply Polyak averaging with exponential decay rate $0.99$ when evaluating the test accuracy in all three algorithms. 
The common practice of training deep CNNs using custom-tuned learning rate decay schedules does not straightforwardly extend to the second-order case. However, \cite{grosse2016kronecker} suggests that Polyak averaging \citep{polyak1992acceleration} can obviate the need for learning rate decay while still achieving high test accuracy. In order to ensure a fair comparison, we apply Polyak averaging with exponential decay rate $0.99$ when evaluating the test accuracy for all three algorithms.
% To make the HF/blockwise HF methods work more stable, we apply the "slow start" of CGs: set maximum number of CGs to be 2 for the first 10 updates and the maximum number of CGs to 30. The second-order
% \begin{table}[h]
%   \caption{GPU time consuming per epoch, fix CGs 30}
%   \label{tab:cnntime}
%   \centering
%   \begin{tabular}{l|lll}%{lll|ll|ll}
%     \toprule
%    	 Curvature mini-batch size & 128 & 256 & 512\\
%         \midrule
%       The HF method (s) & 164 & 305 & 556\\
%    	\midrule
% The blockwise HF method (s)  & 195 & 356& 631\\
%          \bottomrule
%   \end{tabular}
% \end{table}

\begin{figure}[t!]
\centering 
\includegraphics[width=5.0in]{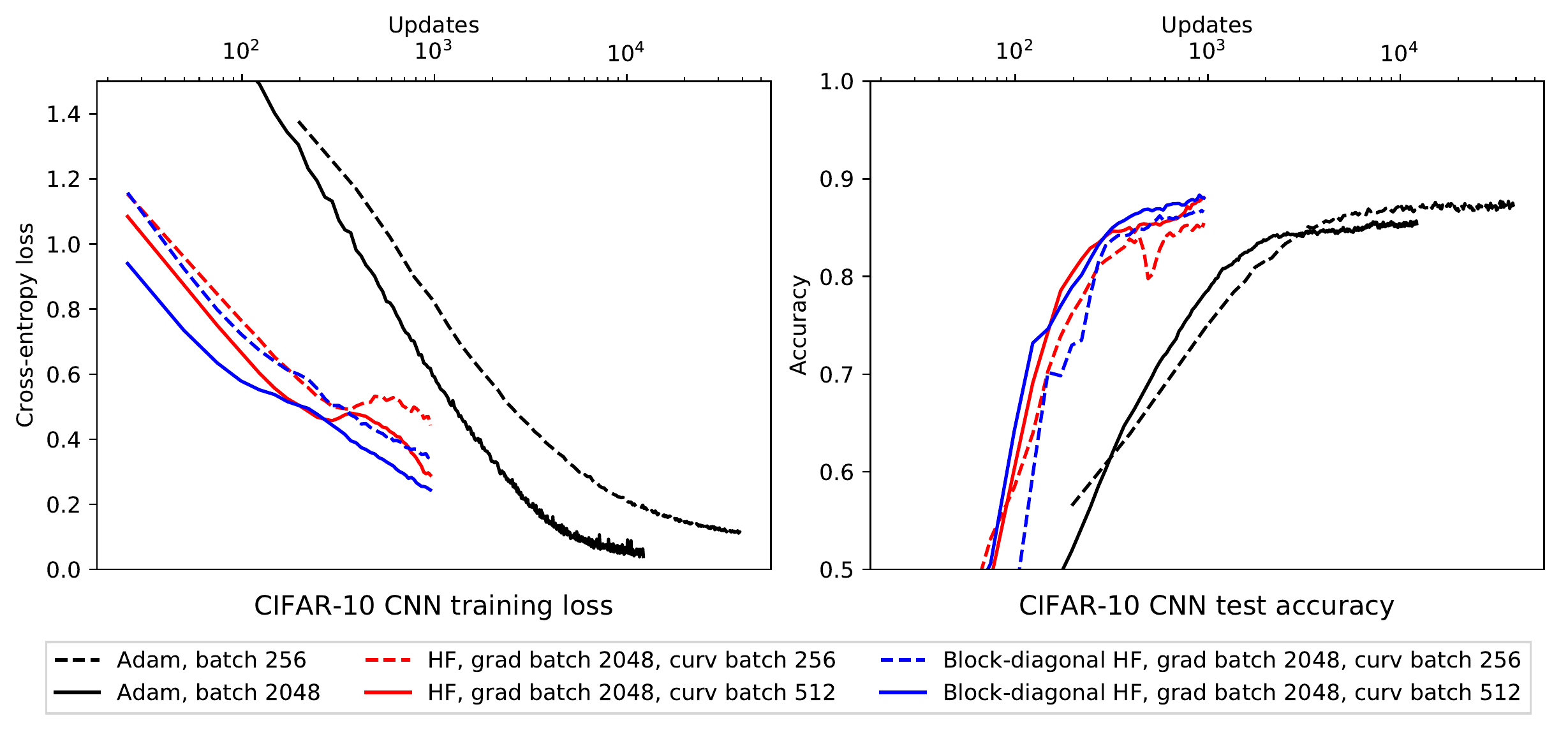}
\vspace{-10pt}
\caption{Performance comparison for a simplified residual CNN on the CIFAR-10 image classification task.  The early epochs of Adam have loss greater than 1.5, while all models are run until the test accuracy decreases or for 1000 epochs.
%The upper two plots compare Adam (with mini-batch sizes 2048 and 256) to block-diagonal HF (with gradient mini-batch 2048 and curvature mini-batch 512). The lower two compare block-diagonal HF to HF, both with gradient mini-batch size 2048 and curvature mini-batch sizes 512 and 256. All plots use a log scale for the number of updates.
}
\label{fig:cnn}
\vspace{-15pt}
\end{figure}

% \begin{figure}[h]
% \centering 
% \includegraphics[width=2.5in]{images/CNNtrainhf.pdf}
% \includegraphics[width=2.5in]{images/CNNtesthf.pdf}

% \caption{Performance comparison on deep CNN network for CIFAR-10 classification task: Blockwise HF  vs. HF. The gradient batch size is $2048$ and curvature mini-batch sizes are $512, 256, 128$ for both methods.}
% \label{fig:cifar10}
% \end{figure}

A performance comparison between block-diagonal HF, HF, and Adam is found in Figure \ref{fig:lstm}. Similar to the autoencoder case, the block-diagonal method with large mini-batches requires far fewer updates to achieve lower training loss and better test accuracy than Adam with any mini-batch size. Furthermore, compared to the HF method, the block-diagonal HF method requires fewer updates, achieves better minima, and exhibits less performance deterioration for small curvature mini-batch sizes.

A performance comparison between block-diagonal Hessian-free, Hessian-free, and Adam is found in Figure \ref{fig:cnn}. Block-diagonal HF with large mini-batches obtains comparable test accuracy to Adam with small ones. Furthermore, the block-diagonal method achieves slightly better training loss and higher test accuracy---and substantially more stable training---than Hessian-free for three different curvature mini-batch sizes.

Although not plotted in figures, the time consumption of block-diagonal HF and that of the HF are comparable in our experiments. The time per iteration of block-diagonal HF and HF is 5-10 times larger than that of the Adam method. However, the total number of iterations of block-diagonal HF and HF are much smaller than Adam and they have potential benefit of parallelization for large mini-batches.
% {Move to experiment} One way to avoid possible harm induced by these inaccuracy is to use short CG iterations. Another way to activate remove the error based on the network architecture. Deep neural networks are built of layers and blocks. It is reasonably to assume that inter correlation between different blocks are much weaker than correlation intra each block. Hence, we propose a blockwise Hessian free method, which only takes into account the block diagonal curvature information. 

% The CG iteration is terminated when the residual r is sufficiently small, or when a prescribed number of CG iterations have been performed.

% \cite{keskar2017large}  Increasing batch size leads to a loss in generalization performance. The generalization gap is correlated with a marked shapness of the minimizers obtained by large-batch methods. ``
% The number of CGs and the mini-batch sizes determines the performance. Larger curvature-vector batch size needs larger number of CGs, while smaller curvature-vector batch size can only tolerate smaller number of CGs. 

\vspace{-3pt}
\section{Conclusion and Discussion}\label{sec:conclusion}
\vspace{-2pt}
We propose a block-diagonal HF method for training neural networks. This approach divides network parameters into blocks, then separates the conjugate gradient subproblem independent for each parameter block. This extension to the original HF method reduces the number of updates needed for training several deep learning models while improving training stability and reaching better minima. Compared to first-order methods including the popular Adam optimizer, block-diagonal HF scales significantly better to large mini-batches, requiring an order of magnitude fewer updates in the large-batch regime.

Our results strengthen the claim of \cite{collobert2004large} that ``the more
block-diagonal the Hessian, the easier it is to train'' a neural network by showing that, in the case of Hessian-free optimization, simply ignoring off-block-diagonal curvature terms improves convergence properties.

Due to the separability of the subproblems for different parameter blocks, the block-diagonal HF method we introduce is inherently more parallelizable than the ordinary HF method. Future work can take advantage of this feature to apply the block-diagonal HF method to large-scale machine learning problems in a distributed setting.

% In this paper, we study the second-order methods for training the deep neural network and propose a blockwise HF method which improves the performance of the original HF method for various network architectures. One attractive feature of the \emph{blockwise HF/HF} method is the ability of handling large batch size and hence would be very efficient in the distributed computing scenario. Our future direction is to explore the benefit of paralleling the blockwise HF method  to deal with very large scale learning problems.

%\input{numerical}

% \section*{References}
%References follow the acknowledgments. Use unnumbered first-level
%heading for the references. Any choice of citation style is acceptable
%as long as you are consistent. It is permissible to reduce the font
%size to \verb+small+ (9 point) when listing the references. {\bf
%  Remember that you can use a ninth page as long as it contains
%  \emph{only} cited references.}
\medskip
\small{
\bibliography{optimization}
\bibliographystyle{abbrvnat}
}
\clearpage

\onecolumn
\appendix

\end{document}